\documentclass{article}
\usepackage[utf8]{inputenc}
\usepackage{macros}

\usepackage{graphicx}
\graphicspath{ {./img/} }

\newcommand{\OPT}{\mathsf{OPT}}

\DeclareMathOperator{\softmax}{softmax}

\newcommand{\st}{\mathrm{s.t.}}

\newcommand{\group}{\mathrm{group}}

\usepackage{caption}
\usepackage{subcaption}
\usepackage[capitalize]{cleveref}
\usepackage{todonotes}
\usepackage{sidecap}

\title{Performance of $\ell_1$ Regularization for Sparse Convex Optimization}
\author{
Kyriakos Axiotis \\ Google Research \\ \texttt{axiotis@google.com} \and
Taisuke Yasuda\thanks{This work was done while T.Y. was an intern at Google Research.} \\ Carnegie Mellon University \\ \texttt{taisukey@cs.cmu.edu}
}
\date{}

\begin{document}

\maketitle
\thispagestyle{empty}

\begin{abstract}
Despite widespread adoption in practice, guarantees for the LASSO and Group LASSO are strikingly lacking in settings beyond statistical problems, and these algorithms are usually considered to be a heuristic in the context of sparse convex optimization on deterministic inputs. We give the first recovery guarantees for the Group LASSO for sparse convex optimization with vector-valued features. We show that if a sufficiently large Group LASSO regularization is applied when minimizing a strictly convex function $l$, then the minimizer is a sparse vector supported on vector-valued features with the largest $\ell_2$ norm of the gradient. Thus, repeating this procedure selects the same set of features as the Orthogonal Matching Pursuit algorithm, which admits recovery guarantees for any function $l$ with restricted strong convexity and smoothness via weak submodularity arguments. This answers open questions of Tibshirani et al. \cite{TBFHSTT2012} and Yasuda et al. \cite{YBCFFM2023}. Our result is the first to theoretically explain the empirical success of the Group LASSO for convex functions under general input instances assuming only restricted strong convexity and smoothness. Our result also generalizes provable guarantees for the Sequential Attention algorithm, which is a feature selection algorithm inspired by the attention mechanism proposed by Yasuda et al. \cite{YBCFFM2023}.

As an application of our result, we give new results for the column subset selection problem, which is well-studied when the loss is the Frobenius norm or other entrywise matrix losses. We give the first result for general loss functions for this problem that requires only restricted strong convexity and smoothness.
\end{abstract}

\clearpage
\setcounter{page}{1}

\section{Introduction}

A common task in modern machine learning is to sparsify a large model by selecting a subset of its inputs. This often leads to a number of improvements to the model such as interpretability and computational efficiency due to the smaller size of the model, as well as improved generalizability due to removal of noisy or redundant features. For these reasons, feature selection and sparse optimization is a heavily studied subject in signal processing, statistics, machine learning, and theoretical computer science. We continue this line of investigation by studying the following sparse optimization problem \cite{SSZ2010, LS2017, EKDN2018}: design an efficient algorithm such that, given $l:\mathbb R^n\to\mathbb R$ and a sparsity parameter $k$, outputs a sparse solution $\tilde\bfbeta$ such that
\begin{equation}\label{eq:sparse-optimization}
    l(\bfzero) - l(\tilde\bfbeta) \geq \gamma \parens*{l(\bfzero) - \min_{\bfbeta\in\mathbb R^n : \norm{\bfbeta}_0 \leq k} l(\bfbeta)}, \qquad \norm{\bfbeta}_0 \coloneqq \abs*{\braces*{i\in[n] : \bfbeta_i \neq 0}}
\end{equation}
for some approximation factor $\gamma > 0$. In practice, there is also much interest in feature selection for vector-valued features, due to a widespread usage of vector representations of discrete features via embeddings \cite{SWY1975,WDLSA2009,PSM2014, PRPG2022}, as well as for applications to block sparsification for hardware efficiency \cite{NUD2017, RPYU2018}, structured sparsification when pruning neurons in neural nets \cite{AS2016, SCHU2017} or channels and filters in convolutional nets \cite{LL2016, WWWCL2016, LKDSG2017, MMK2020, MK2020}. In such vector-valued or group settings, the $n$ inputs $\bfbeta\in\mathbb R^n$ are partitioned into $t$ disjoint groups of features $T_1, T_2, \dots, T_t\subseteq[n]$, and we would like to select whole groups of features at a time. We thus also study the question of solving
\begin{equation}\label{eq:group-sparse-optimization}
    l(\bfzero) - l(\tilde\bfbeta) \geq \gamma \parens*{l(\bfzero) - \min_{\bfbeta\in\mathbb R^n : \norm{\bfbeta}_\group \leq k} l(\bfbeta)}, \qquad \norm{\bfbeta}_\group \coloneqq \abs*{\braces*{i\in[t] : \bfbeta\vert_{T_i} \neq 0}}
\end{equation}
where $\bfbeta\vert_{T_i}$ denotes the $\abs{T_i}$-dimensional vector obtained by restricting $\bfbeta$ to the coordinates $j\in T_i$.\footnote{We also allow for $\bfbeta\vert_{T_i}$ to denote the corresponding $n$-dimensional vector padded with zeros outside of $T_i$ whenever this makes sense.}

Although problems \eqref{eq:sparse-optimization} and \eqref{eq:group-sparse-optimization} are computationally challenging problems in general \cite{Nat1995, FKT2015, GV2021, PSZ2022}, a multitude of highly efficient algorithms have been proposed for solving these problems in practice. Perhaps one of the most popular algorithms in practice is the use of $\ell_1$ regularization. That is, if we wish to optimize a function $l:\mathbb R^n\to\mathbb R$ over $k$-sparse inputs $\braces{\bfbeta\in\mathbb R^n : \norm{\bfbeta}_0 \leq k}$, then we instead optimize the $\ell_1$-regularized objective
\begin{equation}\label{eq:lasso}
    \min_{\bfbeta\in\mathbb R^n} l(\bfbeta) + \lambda\norm{\bfbeta}_1.
\end{equation}
The resulting optimal solution $\bfbeta^*$ often has few nonzero entries and thus helps identify a sparse solution. This idea was first introduced for the linear regression problem by Tibshirani \cite{Tib1996}, known as the \emph{LASSO} in this case, and has subsequently enjoyed wide adoption in practice in applications far beyond the original scope of linear regression. For the group sparsification setting, one can consider a generalization of the LASSO known as the \emph{Group LASSO} \cite{Bak1999, YL2006}, which involves minimizing the following objective:
\begin{equation}\label{eq:group-lasso}
    \min_{\bfbeta\in\mathbb R^n} l(\bfbeta) + \lambda\sum_{i=1}^t \norm{\bfbeta\vert_{T_i}}_2
\end{equation}
That is, the regularizer is now the sum of the $\ell_2$ norms of each group of variables $T_i$ for $i\in[t]$. In practice, this encourages groups of variables to be selected at a time, which facilitates feature selection in the group setting. We refer the reader to the monograph \cite{HTW2015} on the LASSO and its generalizations for further references and discussion.

\subsection{Related work: prior guarantees for \texorpdfstring{$\ell_1$}{L1} regularization}

Due to the practical importance of solving \eqref{eq:lasso} and \eqref{eq:group-lasso}, there has been an intense focus on theoretical work surrounding these optimization problems, especially for the \emph{sparse linear regression} problem, i.e., when $l(\bfbeta) = \norm{\bfX\bfbeta-\bfy}_2^2$ is the least squares objective for a design matrix $\bfX$ and target vector $\bfy$. However, as remarked in a number of works \cite{FKT2015, GV2021, PSZ2022}, recovery guarantees for the LASSO and the Group LASSO are strikingly lacking in settings beyond statistical problems with average-case inputs or strong assumptions on the input, and is usually considered to be a heuristic in the context of sparse convex optimization for deterministic inputs. For example, one line of work focuses on the linear regression problem in the setting where the target vector $\bfy$ is exactly a $k$-sparse linear combination $\bfX\bfbeta$ for some $\norm{\bfbeta}_0 \leq k$ plus i.i.d.\ Gaussian noise, and we seek guarantees on the solution to \eqref{eq:lasso} \cite{DS1989, CDS1998, Tro2006, CRT2006, CT2007, Can2008, BRT2009, Zho2009, RWY2010, BCFS2014} when $\bfX$ satisfies the \emph{restricted isometry property (RIP)} or its various relaxations such as the \emph{restricted eigenvalue condition (RE)}. This can be viewed as an instantiation of \eqref{eq:sparse-optimization} for $l(\bfbeta) = \norm{\bfX\bfbeta-\bfy}_2^2$, under the assumption that there exists an approximate global optimum of $l$ that is exactly $k$-sparse. Statistical consistency results have also been established, which also assume a ``true'' $k$-sparse target solution \cite{ZY2006, MGB2008}. A more recent line of work has studied algorithms for sparse linear regression problem under a correlated Gaussian design matrix with other general structural assumptions on the covariance matrix \cite{KKMR2021, KKMR2022, KKMR2023}. All of these works exclude the consideration of worst-case error on a desired $k$-sparse target solution, which is an undesirable restrictive assumption when solving \eqref{eq:sparse-optimization} in general. Indeed, one of the most remarkable aspects about the LASSO is its empirical success on a wide variety of real input distributions that can be far from Gaussian or even general i.i.d.\ designs. Thus, gaining a theoretical explanation of the success of the LASSO in more general settings is a critical question in this literature. 

\begin{Question}
\label{q:input-distribution}
Why are the LASSO and Group LASSO successful on general input distributions, beyond statistical settings?
\end{Question}

An important exception is the work of \cite{YBCFFM2023}, which establishes that in the setting of sparse linear regression, a sequential variation on the LASSO known as the Sequential LASSO \cite{LC2014}, in which the LASSO is applied sequentially $k$ times to select $k$ inputs one at a time, is in fact equivalent to the Orthogonal Matching Pursuit algorithm (OMP) \cite{PRK1993, Tro2004}.\footnote{We also note a work of \cite{TBFHSTT2012}, which proposes a similar procedure called the Strong Sequential Rule of sequentially zeroing out variables using the LASSO, but does not obtain provable guarantees for the resulting selected features.} The work of \cite{DK2011} showed that OMP achieves bounds of the form of \eqref{eq:sparse-optimization} whenever $\bfX$ satisfies a restricted isometry property in the absence of additional distributional assumptions on the input instance. From \cite{YBCFFM2023}, it follows that the Sequential LASSO does as well. Thus, the works of \cite{DK2011, YBCFFM2023} provide a form of an answer to Question \ref{q:input-distribution} for the sparse linear regression problem, for general inputs with RIP.

Given the previous success of analyzing the LASSO for general inputs under RIP, one may ask for generalizations of this result to other objective functions, such as generalized linear models, logistic regression, or even general sparse convex optimization. Indeed, as mentioned previously, the LASSO and Group LASSO are used in practice in settings far beyond linear regression, and fast algorithms for solving the optimization problems of \eqref{eq:lasso} and \eqref{eq:group-lasso} are plentiful in the literature \cite{LLAN2006, KKB2007, SFR2007, MGB2008, HTF2009, FHT2010, BPCPE2011, BJMO+2011, HMR2023}. However, none of these works provide satisfactory answers on \emph{why} the LASSO and Group LASSO are successful at selecting a good sparse set of inputs.

\begin{Question}
\label{q:convex-objective}
Why are the LASSO and Group LASSO successful on general convex objectives, beyond $\ell_2$ linear regression? Why do they select a sparse set of inputs? Which inputs are chosen?
\end{Question}

While the work of \cite{YBCFFM2023} provides answers for the sparse linear regression problem by showing that the selected inputs are precisely the inputs selected by OMP, their analysis relies on specific geometric properties of the linear regression loss such as the Pythagorean theorem and the fact that the dual of the LASSO objective is a Euclidean norm projection onto a polytope \cite{OPT2000, TT2011}, and thus the techniques there do not immediately generalize even to specific problems such as $\ell_p$ regression or regularized logistic regression. Such a generalization is left as a central open question in their work. Similarly, the work of \cite{TBFHSTT2012} asks the question of why sequentially discarding variables using the LASSO performs so well.

\subsection{Our results}

The main result of this work is a resolution of Question \ref{q:convex-objective} for both the LASSO \eqref{eq:lasso} and the Group LASSO \eqref{eq:group-lasso} setting for any strictly convex objective function $l$. To state our results, we first recall the (Group) Sequential LASSO and (Group) OMP algorithms in Algorithms \ref{alg:group-sequential-lasso} and \ref{alg:group-omp}, which are both iterative algorithms that maintain a set of selected features $S\subseteq[t]$ by adding one feature at a time starting with $S = \varnothing$.

\begin{algorithm}
\caption{Group Sequential LASSO.}
\label{alg:group-sequential-lasso}
\begin{algorithmic}[1]
  \Function{GroupSequentialLASSO}{objective $l$, sparsity $k$, iterations $k'$}
  \State Initialize $S \gets \varnothing$
  \For{$r=1$ to $k'$}
    \State Let $\tau \coloneqq \sup\braces{\lambda > 0 : \exists i\in \overline S, \bfbeta^\lambda\vert_{T_i} \neq 0}$ for
    \[
        \bfbeta^\lambda \coloneqq \arg\min_{\bfbeta\in\mathbb R^n} l(\bfbeta) + \lambda \sum_{i\in\overline S} \norm{\bfbeta\vert_{T_i}}_2
    \]
    \State For $\eps>0$ sufficiently small, let $i^*\in\overline S$ be such that $\bfbeta^{\tau-\eps}\vert_{T_{i^*}} \neq 0$
    \State Update $S \gets S \cup \{i^*\}$
  \EndFor
  \State \Return $S$
  \EndFunction
\end{algorithmic}
\end{algorithm}

\begin{algorithm}
\caption{Group Orthogonal Matching Pursuit.}
\label{alg:group-omp}
\begin{algorithmic}[1]
  \Function{GroupOMP}{objective $l$, sparsity $k$, iterations $k'$}
  \State Initialize $S \gets \varnothing$
  \For{$r=1$ to $k'$}
    \State Let
    \[
        \bfbeta^\infty \coloneqq \arg\min_{\substack{\bfbeta\in\mathbb R^n \\ \forall i\in\overline S, \bfbeta\vert_{T_i} = 0}} l(\bfbeta)
    \] \label{line:optimize-beta}
    \State Let $i^*\in\overline S$ be such that $\norm{\nabla l(\bfbeta^\infty)\vert_{T_{i^*}}}_2^2 = \max_{i\in\overline S}\norm{\nabla l(\bfbeta^\infty)\vert_{T_i}}_2^2$ \label{line:select-gradient}
    \State Update $S \gets S \cup \{i^*\}$
  \EndFor
  \State \Return $S$
  \EndFunction
\end{algorithmic}
\end{algorithm}

We show that the result of \cite{YBCFFM2023} generalizes to the setting of group-sparse convex optimization: the Group Sequential LASSO update rule selects a group of features $T_i\subseteq[n]$ that maximizes the $\ell_2$ gradient mass $\norm{\nabla l(\bfbeta)\vert_{T_i}}_2^2$, i.e., the same update rule as Group OMP. Our analysis simultaneously gives a substantial simplification as well as a generalization of the analysis of \cite{YBCFFM2023}, which gives us the flexibility to handle both group settings as well as general convex functions.

\begin{Theorem}
\label{thm:sequential-lasso-vs-omp}
Let $l:\mathbb R^n\to\mathbb R$ be strictly convex. Let $S\subseteq[t]$ be a set of currently selected features. For each $\lambda>0$, define
\[
    \bfbeta^\lambda \coloneqq \arg\min_{\bfbeta\in\mathbb R^n} l(\bfbeta) + \lambda \sum_{i\in\overline S} \norm{\bfbeta\vert_{T_i}}_2
\]
and let $\tau \coloneqq \sup\braces{\lambda > 0: \exists i\in \overline S, \bfbeta^\lambda\vert_{T_i} \neq 0}$ and let $\bfbeta^\infty \coloneqq \bfbeta^\tau = \lim_{\lambda\to\infty} \bfbeta^\lambda$. Then for $\lambda = \tau - \eps$ for all $\eps>0$ sufficiently small, $\bfbeta^\lambda\vert_{T_i} \neq 0$ only if $\norm{\nabla l(\bfbeta^\infty)\vert_{T_i}}_2^2 = \max_{j\in\overline S} \norm{\nabla l(\bfbeta^\infty)\vert_{T_j}}_2^2$. 
\end{Theorem}
\begin{proof}
We give our discussion of this result in Section \ref{sec:sequential-lasso-vs-omp}.
\end{proof}

In other words, if we add the Group LASSO regularization only on unselected features $i\in\overline S$ and take $\lambda$ as large as possible without causing the solution $\bfbeta^\lambda$ to be zero, then $\bfbeta^\lambda$ must be supported on a group of features maximizing the $\ell_2$ gradient mass at $\bfbeta^\infty$ among the unselected features $i\in\overline S$. Furthermore, note that $\bfbeta^\infty$ is exactly the minimizer of $l(\bfbeta)$ subject to the constraint that $\bfbeta\vert_{T_i} = 0$ for every $i\in \overline S$. Thus, in the non-group setting, this algorithm sequentially selects a feature $i\in[n]$ that maximizes $\abs{\nabla l(\bfbeta^\infty)_i}$, which is exactly the OMP update rule analyzed in \cite{SSZ2010, LS2017, EKDN2018}. The works of \cite{SSZ2010, LS2017, EKDN2018} show that this OMP update rule gives a guarantee of the form of \eqref{eq:sparse-optimization} with an approximation factor $\gamma$ depending on the \emph{restricted strong convexity (RSC)} of $l$, which is a generalization of the RIP parameter for matrices to general functions. Thus, as reasoned in \cite{YBCFFM2023}, the Sequential LASSO for general functions $l$ inherits this guarantee of OMP. We also show in Section \ref{sec:group-omp} that the group version of the OMP update rule obtained here based on selecting the group with the largest $\ell_2$ gradient mass $\norm{\nabla l(\bfbeta)\vert_{T_i}}_2^2$ in fact also gives an analogous guarantee. In particular, we give guarantees for Group OMP both in the setting of outputting exactly $k$-group-sparse solutions (Corollary \ref{cor:k-sparse}) as well as bicriteria solutions that use a slightly larger sparsity to get within an additive $\eps$ of the function value of the optimal $k$-sparse solution (Corollary \ref{cor:bicriteria}), restated below. 

\begin{restatable*}[Exactly $k$-group-sparse solutions]{Corollary}{CorSparseOMP}
\label{cor:k-sparse}
After $k$ iterations of Algorithm \ref{alg:group-omp}, $\bfbeta^\infty$ (Line \ref{line:optimize-beta}) has group sparsity $\norm{\bfbeta^\infty}_\group \leq k$ and satisfies \eqref{eq:group-sparse-optimization} with
\[
    \gamma = 1-\exp\parens*{-\frac{\mu_{2k}}{L_1}}\,,
\]
where $\mu_{2k}$ is a lower bound on the restricted strong
convexity constant of $l$ at group sparsity $2k$ and $L_1$ is an upper bound on the restricted smoothness constant of $l$ at group sparsity $1$ (see Definition \ref{def:rsc}).
\end{restatable*}

\begin{restatable*}[Bicriteria sparsity with $\eps$ additive error]{Corollary}{CorBicriteriaOMP}
\label{cor:bicriteria}
After $k'$ iterations of Algorithm \ref{alg:group-omp}, for
\[
    k' \geq 
    k\cdot \frac{L_1}{\mu_{k+k'}}\log\frac{l(\bfbeta^{(0)}) - l(\bfbeta^*)}{\eps},
\]
then $\bfbeta^\infty$ (Line \ref{line:optimize-beta}) has group sparsity $\norm{\bfbeta^\infty}_\group \leq k'$ and satisfies
\[
    l(\bfbeta^\infty) \leq l(\bfbeta^*) + \eps\,,
\]
where $\mu_{k+k'}$ is a lower bound on the restricted strong
convexity constant of $l$ at group sparsity $k+k'$ and $L_1$ is an upper bound on the restricted smoothness constant of $l$ at group sparsity $1$ (see Definition \ref{def:rsc}).
\end{restatable*}

We additionally note that our analysis also immediately extends to an analysis of a local search version of OMP, known as OMP with Replacement (Algorithm \ref{alg:group-ompr}) \cite{JTD2011, AS2020}, which gives a bicriteria sparsity bound which does not depend on $\eps$ (Corollary \ref{cor:ompr-bicriteria}). 

\begin{restatable*}[Bicriteria sparsity with $\eps$ additive error]{Corollary}{CorBicriteriaOMPR}
\label{cor:ompr-bicriteria}
After $R$ iterations of Algorithm \ref{alg:group-ompr} with $k'\geq k\parens*{\frac{L_2^2}{\mu_{k+k'}^2} + 1}$, for
\[
    R \geq 
    k\cdot \frac{L_2}{\mu_{k+k'}}\log\frac{l(\bfbeta^{(0)}) - l(\bfbeta^*)}{\eps},
\]
then $\bfbeta^\infty$ (Line \ref{line:optimize-beta-ompr}) has group sparsity $\norm{\bfbeta^\infty}_\group \leq k'$ and satisfies
\[
    l(\bfbeta^\infty) \leq l(\bfbeta^*) + \eps\,,
\]
where $\mu_{k+k'}$ is a lower bound on the restricted strong
convexity constant of $l$ at group sparsity $k+k'$ and $L_2$ is an upper bound on the restricted smoothness constant of $l$ at group sparsity $2$ (see Definition \ref{def:rsc}).
\end{restatable*}

This variant of OMP can be analogously simulated by the LASSO as well, leading to a new LASSO-based feature selection algorithm which we call (Group) Sequential LASSO with Replacement.

\subsubsection{Techniques}

Our main technique involves exploiting the correspondence between variables of a primal optimization problem with the gradient of the dual optimization problem, via the \emph{Fenchel--Young inequality} (Theorem \ref{thm:fenchel-young}).

We start with an observation given by \cite{GVR2010}. When we take the dual of the LASSO objective, then the resulting problem involves minimizing the \emph{Fenchel dual} $l^*$ of $l$ (Definition \ref{def:fenchel-conjugate}), subject to a hypercube constraint set. When the regularization $\lambda$ is sufficiently large (say larger than some threshold $\tau$), then this increases the size of the constraint set large enough to contain the global minimizer of the Fenchel dual $l^*$, and thus the gradient of $l^*$ vanishes at this minimizer. Then by the equality case of the Fenchel--Young inequality, this implies that the corresponding primal variable $\bfbeta$ is zero as well. On the other hand, if $\lambda$ is smaller than this threshold point $\tau$, only some coordinates will be unconstrained (i.e. strictly feasible), while others coordinates will become constrained by the smaller $\lambda$. In this case, the strictly feasible coordinates will have zero gradient, which leads to zeroes in the corresponding primal variable $\bfbeta$ and thus a sparse solution. The argument until this point is known in prior work, and \cite{GVR2010} used this observation to give an algorithm which tunes the value of $\lambda$ such that at least $k$ variables are selected in a single application, while \cite{TBFHSTT2012} proposed a sequential procedure with better empirical performance.

Our central observation, inspired by the work of \cite{YBCFFM2023}, is that if we regularize strongly enough such that only one feature is selected at a time via the LASSO, then this feature is the one maximizing the absolute value of the gradient. Indeed, note that if $\lambda$ is just slightly smaller than the threshold point $\tau$, then the global minimizer $\bfu^*\in\mathbb R^n$ of $l^*$ just slightly violates exactly a single constraint in the dual problem, which corresponds to the feature $i^*\in[n]$ with the largest absolute coordinate value $\abs{\bfu^*_i}$ in the dual variable. We show that for such $\lambda$, all other coordinates $j\in[n]\setminus\{i^*\}$ are unconstrained optimizers and thus the gradient is $\bfzero$ (Lemma \ref{lem:dual-solution-gradient}). Thus, by the equality case of the Fenchel--Young inequality, this corresponds to a primal variable $\bfbeta$ that is supported only on this coordinate $i^*\in[n]$. The crucial next step then is to \emph{apply the Fenchel--Young inequality again in the dual direction}: via the Fenchel--Young inequality, this coordinate $i^*\in[n]$ maximizes the absolute coordinate value of the dual variable $\bfu$, and thus is the coordinate that maximizes the absolute coordinate value of the gradient of the primal variable $\bfbeta$. Thus, this selects a coordinate which follows the first step of the OMP update rule. While we have sketched the proof only for this first step in the non-group setting, the analysis also carries through for all steps of the OMP algorithm, as well as for the group setting. Thus, this establishes the equivalence between (Group) Sequential LASSO and (Group) OMP for general convex functions.

\subsubsection{Connections to analysis of attention mechanisms}

As noted in \cite{YBCFFM2023}, we make a connection of our work to the analysis of recently popularized techniques for discrete optimization via continuous and differentiable relaxations inspired by the \emph{attention mechanism} \cite{VSPUJGKP2017}. The attention mechanism can be viewed as a particular algorithm for the sparse optimization problem \eqref{eq:sparse-optimization}, in which an additional set of variables $\bfw\in\mathbb R^n$ are introduced, and we solve a new optimization problem
\begin{equation}\label{eq:softmax-attention}
    \min_{\bfw,\bfbeta\in\mathbb R^n} l(\softmax(\bfw)\odot\bfbeta),
\end{equation}
where $\odot$ denotes the Hadamard (entrywise) product and $\softmax(\bfw)\in\mathbb R^n$ is defined as
\[
    \softmax(\bfw)_i \coloneqq \frac{\exp(\bfw_i)}{\sum_{j=1}^n \exp(\bfw_j)}.
\]
The idea is that $\bfw$ serves as a measure of ``importance'' of each feature $i\in[n]$, and the softmax allows for a differentiable relaxation for the operation of selecting the most ``important'' feature when minimizing the loss $l$. Alternatively, $\bfw$ can be viewed as the amount of ``attention'' placed on feature $i\in[n]$ by the algorithm. Such ideas have been applied extremely widely in machine learning, with applications to feature selection \cite{LLY2021, YBCFFM2023}, feature attribution \cite{AP2021}, permutation learning \cite{MBLS2018}, neural architecture search \cite{LSY2019}, and differentiable programming \cite{NLS2015}. Thus, it is a critical problem to obtain a theoretical understanding of subset selection algorithms of the form of \eqref{eq:softmax-attention}. 

The work of \cite{YBCFFM2023} showed that a slight variation on \eqref{eq:softmax-attention} is in fact amenable to analysis when $l$ is the problem of least squares linear regression. In this case, \cite{YBCFFM2023} show (using a result of \cite{Hof2017}) that if we instead consider
\begin{equation}\label{eq:linear-attention}
    \min_{\bfw,\bfbeta\in\mathbb R^n} l(\bfw\odot\bfbeta) + \frac{\lambda}{2}\parens*{\norm{\bfw}_2^2 + \norm{\bfbeta}_2^2}
\end{equation}
i.e., remove the softmax and add $\ell_2$ regularization, then this is in fact equivalent to the $\ell_1$-regularized problem considered in \eqref{eq:lasso}. In Lemma \ref{lem:group-hoff}, we show a generalization of this fact to the group setting, by showing that if we have $t$ features corresponding to disjoint subsets of coordinates $T_1, T_2, \dots, T_t\subseteq[n]$, then multiplying each of the features $\bfbeta\vert_{T_i}$ by a single ``attention weight'' $\bfw_i$ for $\bfw\in\mathbb R^t$ gives a similar correspondence to the Group LASSO algorithm \eqref{eq:group-lasso}. Thus, the attention-inspired feature selection algorithm given in Algorithm \ref{alg:group-sequential-attention} also enjoys the same guarantees as the Group Sequential LASSO algorithm. We note that this generalization to the group setting is particularly important for the various applications in attention-based subset selection algorithms, due to the fact that the objects $\bfbeta\vert_{T_i}$ being selected are often large vectors in these applications.

\begin{algorithm}
\caption{Group Sequential Attention.}
\label{alg:group-sequential-attention}
\begin{algorithmic}[1]
  \Function{GroupSequentialAttention}{objective $l$, sparsity $k$, iterations $k'$}
  \State Initialize $S \gets \varnothing$
  \For{$r=1$ to $k'$}
    \State Let $\tau \coloneqq \sup\braces{\lambda > 0 : \exists i\in \overline S, \bfbeta^\lambda\vert_{T_i} \neq 0}$ for
    \[
        \bfbeta^\lambda \coloneqq \arg\min_{\bfw\in\mathbb R^t, \bfbeta\in\mathbb R^n} l(\bfbeta_\bfw) + \frac{\lambda}{2} \sum_{i\in\overline S} \bfw_i^2 + \norm{\bfbeta\vert_{T_i}}_2^2, \qquad \bfbeta_\bfw\vert_{T_i} \coloneqq  \bfw_i \cdot \bfbeta\vert_{T_i}
    \]
    \State For $\eps>0$ sufficiently small, let $i^*\in\overline S$ be such that $\bfbeta^{\tau-\eps}\vert_{T_{i^*}} \neq 0$
    \State Update $S \gets S \cup \{i^*\}$
  \EndFor
  \State \Return $S$
  \EndFunction
\end{algorithmic}
\end{algorithm}

Finally, we also note that our analysis of Hadamard product-type of algorithms of the form of \eqref{eq:linear-attention} may prove to be useful in the analysis of similar algorithms in the literature of online convex optimization that have been developed to solve sparse optimization problems \cite{AW2020a, AW2020b, Chi2022}.

\subsubsection{Applications to column subset selection}

As a corollary of our analyses of group feature selection algorithms, we obtain the first algorithms for the \emph{column subset selection (CSS)} problem for general loss functions with restricted strong convexity and smoothness.

In the CSS problem, we are given an input matrix $\bfX\in\mathbb R^{n\times d}$, and the goal is to select a small subset of $k$ columns $S\subseteq[d]$ of $\bfX$ that minimizes the reconstruction error
\begin{equation}\label{eq:reconstruction-error}
    \min_{\bfV\in\mathbb R^{k\times d}}\norm*{\bfX - \bfX\vert^S \bfV}_F^2,
\end{equation}
where $\bfX\vert^S \in\mathbb R^{n\times k}$ is the matrix $\bfX$ restricted to the columns indexed by $S$. As with sparse linear regression, this problem is known to be computationally difficult \cite{Civ2014}, and thus most works focus on approximation algorithms and bicriteria guarantees to obtain tractable results.

The CSS problem can be viewed as an unsupervised analogue of sparse convex optimization, and has been studied extensively in prior work. In particular, the works of \cite{FGK2011, CM2012, FGK2013, FEGK2015, SVW2015, ABFMRZ2016, LS2017} gave analyses of greedy algorithms for this problem, showing that iteratively selecting columns that maximizes the improvement in reconstruction error \eqref{eq:reconstruction-error} leads to bicriteria sparsity algorithms that depend on the sparse condition number of $\bfX$. In a separate line of work, randomized methods have been employed in the randomized numerical linear algebra literature to sample columns of $\bfX$ that span a good low rank approximation \cite{DV2006, DMM2008, BMD2009, DR2010, BDM2011, CEMMP2015, BW2017, CMM2017}. Furthermore, there has recently been a large body of work aimed at generalizing CSS results to more general loss functions beyond the Frobenius norm, including $\ell_p$ norms \cite{SWZ2017, CGKLPW2017, DWZZR2019,  SWZ2019b, JLLMW2021, MW2021} and other entrywise losses \cite{SWZ2019a, WY2023}. All of these works use complicated arguments and rely heavily on the entrywise structure of the loss function.

We show that by a surprisingly simple argument, we can immediately obtain the first results on column subset selection for general convex loss functions with restricted strong convexity and smoothness. Our key insight is to view this problem not as a column subset selection problem for $\bfX$, but rather a \emph{row subset selection problem for $\bfV$}. That is, note that
\[
    \min_{\abs{S}\leq k}\min_{\bfV\in\mathbb R^{k\times d}}l\parens*{\bfX - \bfX\vert^S \bfV} = \min_{\abs{S}\leq k}\min_{\bfV\in\mathbb R^{d\times d}}l\parens*{\bfX - \bfX\bfV\vert_S}
\]
where $\bfV\vert_S$ zeros out all rows of $\bfV$ not indexed by $S$. Then, this is just a group variable selection problem, where we have $d$ groups given by each of the rows of $\bfV$, and thus we may write this problem as computing
\[
    \OPT = \min_{\bfV\in\mathbb R^{d\times d}, \norm{\bfV}_\group \leq k} l(\bfX - \bfX\bfV)
\]
Thus, by using our guarantees for Group OMP in Corollaries \ref{cor:k-sparse} and \ref{cor:bicriteria} (which also hold for Group Sequential LASSO and Group Sequential Attention by Theorem \ref{thm:sequential-lasso-vs-omp} and Lemma \ref{lem:group-hoff}), we obtain the first algorithm and analysis of the column subset selection problem under general loss functions with restricted strong convexity and smoothness. This gives a substantial generalization of results known in prior work.

\begin{Theorem}[Column subset selection via Group OMP]
Let $\bfX\in\mathbb R^{n\times d}$ and let $l:\mathbb R^{n\times d}\to\mathbb R$ be a strictly convex and differentiable loss function. Let $\bfV\mapsto l(\bfX - \bfX\bfV)$ satisfy $L_1$-group-sparse smoothness and $\mu_{k+k'}$-group-sparse convexity (Definition \ref{def:rsc}), where the groups are the rows of $\bfV$. The following hold: 
\begin{itemize}
    \item Let $\kappa = L_1/\mu_{2k}$. After $k' = k$ iterations, Algorithm \ref{alg:group-omp} outputs a subset $S\subseteq[n]$ of size $\abs{S} \leq k$ such that
    \[
        l(\bfX) - l(\bfX-\bfX\vert^S\bfV) \geq \parens*{1 - e^{-\kappa}}\parens*{l(\bfX) - \OPT}.
    \]
    \item Let $\kappa = L_1/\mu_{k+k'}$. After $k' \geq k\cdot \kappa\log\frac{l(\bfX) - \OPT}{\eps}$ iterations, Algorithm \ref{alg:group-omp} outputs a subset $S\subseteq[n]$ of size $\abs{S} \leq k'$ such that
    \[
        l(\bfX-\bfX\vert^S\bfV) \leq \OPT + \eps.
    \]
\end{itemize}
\end{Theorem}
\begin{proof}
This follows from applying Corollaries \ref{cor:k-sparse} and \ref{cor:bicriteria} to the group-sparse convex optimization formulation of column subset selection.
\end{proof}

Our proof is arguably simpler than prior work even for the Frobenius norm. Indeed, the prior works require arguments that use the special structure of Euclidean projections, whereas we simply observe that CSS is a group-sparse convex optimization problem and use a generalization of techniques for sparse regression. We also immediately obtain analyses for natural algorithms which were previously not considered in the context of column subset selection, such as Group OMP (with Replacement), Group LASSO, and attention-based algorithms. In particular, by applying guarantees for Group OMP with Replacement (Corollary \ref{cor:ompr-bicriteria}), we obtain the first column subset selection algorithm with no dependence on $\eps$ in the sparsity, even for the Frobenius norm problem.

\begin{Theorem}[Column subset selection with Group OMPR]
Let $\bfX\in\mathbb R^{n\times d}$ and let $l:\mathbb R^{n\times d}\to\mathbb R$ be a strictly convex and differentiable loss function. Let $\bfV\mapsto l(\bfX - \bfX\bfV)$ satisfy $L_2$-group-sparse smoothness and $\mu_{k+k'}$-group-sparse convexity (Definition \ref{def:rsc}), where the groups are the rows of $\bfV$. Let $\kappa = L_2/\mu_{k+k'}$ and $k' \geq k(\kappa^2 + 1)$. After $R \geq k\cdot \kappa\log\frac{l(\bfX) - \OPT}{\eps}$ iterations, Algorithm \ref{alg:group-ompr} outputs a subset $S\subseteq[n]$ of size $\abs{S} \leq k'$ such that
\[
    l(\bfX-\bfX\vert^S\bfV) \leq \OPT + \eps.
\]
\end{Theorem}
\begin{proof}
This follows from applying Corollary \ref{cor:ompr-bicriteria} to the group-sparse convex optimization formulation of column subset selection.
\end{proof}

\subsection{Related work: the Forward Stagewise Regression conjecture}

A separate line of work has investigated a closely related connection between the LASSO and OMP-like algorithms. In particular, the ``continuous'' OMP (or coordinate descent) algorithm which updates $\bfbeta^{(t+1)} \gets \bfbeta^{(t)} - \eta\cdot \mathrm{sign}(\nabla_i l(\bfbeta^{(t)}))\bfe_i$ for $i = \arg\max_{i=1}^n \nabla l(\bfbeta^{(t)})$ known as Forward Stagewise Regression is conjectured \cite[Conjecture 2]{RZH2004}) to have the same solution path as the LASSO path (i.e. the set of solutions as $\lambda$ ranges from $0$ to $\infty$) when $\eta\to0$ \cite{EHJT2004, RZH2004, Tib2015, FGM2017}. While a full proof of this conjecture may be useful towards proving our main result, to the best of our knowledge, the only known result towards this conjecture establishes an ``instantaneous'' result which shows the convergence of the difference between the two paths to the gradient \cite[Theorem 1]{RZH2004} under technical assumptions under the underlying loss function such as the monotonicity of the coordinates of the LASSO solution. Our result can be viewed as a full proof of this conjecture in an open ball near $\bfzero$ for general strictly convex differentiable functions, and our techniques may be useful for a full resolution of this conjecture.

\subsection{Related work: algorithms for sparse convex optimization}

While we have argued so far that guarantees for $\ell_1$ regularization in solving \eqref{eq:sparse-optimization} in prior work are limited, other efficient algorithms have in fact been shown to solve \eqref{eq:sparse-optimization}, both for sparse linear regression as well as general sparse convex optimization. Via a connection between convexity and weakly submodular optimization, the works of \cite{SSZ2010, LS2017, EKDN2018} showed that the greedy forward algorithm and Orthogonal Matching Pursuit both give guarantees of the form of \eqref{eq:sparse-optimization}. Efficiency guarantees have
also been given for OMP with Replacement (OMPR)
\cite{SSZ2010,JTD2011,AS2020} and
Iterative Hard Thresholding (IHT) \cite{JTK2014,AS2022}, using the
restricted smoothness and strong convexity properties. Ultimately, these results show
that an $\epsilon$-approximate 
sparse solution can be recovered if we allow
an $O(\kappa)$ blowup to the
sparsity, where $\kappa$ is the restricted
condition number of the problem.

\subsection{Open directions}

We suggest several directions for future study. Our first question is on showing analogous results for the one-shot version of LASSO, which is used much more frequently in practice than the Sequential LASSO. That is, if $\lambda$ is chosen in \eqref{eq:lasso} such that only $k$ nonzero entries are selected, then can we obtain a guarantee of the form of \eqref{eq:sparse-optimization} for this solution? It is known that one-shot variants of OMP or greedy have this type of guarantee \cite{DK2011, EKDN2018} (also called ``oblivious'' algorithms in these works). However, our proof techniques do not immediately apply, since we crucially use the fact that for large enough regularizations $\lambda$, the resulting solution is close to the $\lambda = \infty$ solution, while this is not true when $\lambda$ can be much smaller. 

A second question is whether our results generalize beyond convex functions or not. For example, the analysis of OMP carries through to smooth functions that satisfy the Polyak-Łojasiewicz condition \cite{KNS2016}. Can a similar generalization be shown for our results? There are several parts of our proofs that crucially use convexity, but the LASSO is known to give good results even for nonconvex functions in practice and thus there is still a gap in our understanding of this phenomenon.

Finally, we ask if our analyses for $\ell_1$ regularization can be extended to an analogous result for nuclear norm regularization for rank-constrained convex optimization. In the setting of rank-constrained convex optimization, it has been shown in special cases, such as affine rank minimization, that nuclear norm regularization can be used to efficiently recover low rank solutions \cite{RFP2010}. This suggests that our results may have a natural generalization in this setting as well. In particular, an extension of OMP to the rank-sparse setting was shown by \cite{AS2021}, and thus it is possible that nuclear norm regularization can be used to  simulate this algorithm as well. 

\section{Preliminaries}

Let $l:\mathbb R^n\to\mathbb R$ be strictly convex and differentiable. For each $i\in[t]$, let $T_i\subseteq[n]$ denote the group of variables that belong to the $i$-th feature.

\subsection{Fenchel duality}

We will use the following standard facts about Fenchel duality \cite{BV2004}. 

\begin{Definition}[Fenchel dual]
\label{def:fenchel-conjugate}
Let $l:\mathbb R^n\to\mathbb R$. Then, the Fenchel dual $l^*$ of $l$ is
\[
    l^*(\bfu) \coloneqq \sup_{\bfz\in\mathbb R^n} \bfu^\top\bfz - l(\bfz).
\]
\end{Definition}

\begin{Theorem}[Fenchel--Young inequality]
\label{thm:fenchel-young}
Let $l:\mathbb R^n\to\mathbb R$ be convex and differentiable. Then,
\[
    l(\bfz) + l^*(\bfu) \geq \bfu^\top\bfz
\]
with equality if and only if $\bfu = \nabla l(\bfz)$.
\end{Theorem}

\begin{Theorem}[Conjugacy theorem]
\label{thm:fenchel-conjugacy}
Let $l:\mathbb R^n\to\mathbb R$ be convex. Then, $(l^*)^* = l$.
\end{Theorem}

The following is known about the convexity and differentiability of the Fenchel dual.

\begin{Theorem}[Differentiability of dual, Theorem 26.3, \cite{Roc1970}]
Let $l:\mathbb R^n\to\mathbb R$ be strictly convex and differentiable. Then, $l^*$ is strictly convex and differentiable. 
\end{Theorem}

\subsection{Berge's theorem}

We will use a well-known theorem of Berge on the continuity of the argmin for constrained optimization problems with parameterized constraint sets.

Recall that a correspondence $h:\mathbb R\rightrightarrows \mathbb R^n$ is a set-valued function which maps real numbers $\lambda$ to subsets $h(\lambda)\subseteq\mathbb R^n$. A correspondence $h$ is upper hemicontinuous if for every $\lambda\in\mathbb R$ and every open set $G\subseteq\mathbb R^n$ such that $h(\lambda)\subset G$, there is an open set $U\subseteq\mathbb R$ such that $\tau\in U\implies h(\tau)\subset G$.

\begin{Theorem}[Berge's theorem \cite{Ber1963}]
\label{thm:berge}
Let $g:\mathbb R^n\to\mathbb R$ be a continuous function and let $\varphi:\mathbb R\rightrightarrows\mathbb R^n$ be a continuous correspondence that map into compact sets. Consider the correspondence $h:\mathbb R\rightrightarrows \mathbb R^n$ given by
\[
    h(\lambda) = \braces*{\bfu\in\mathbb R^n : g(\bfu) = \min_{\bfu'\in\varphi(\lambda)} g(\bfu')}
\]
Then, $h$ is upper hemicontinuous. 
\end{Theorem}

The following corollary of Theorem \ref{thm:berge} for strictly convex functions is more useful for our purposes.

\begin{Corollary}[Berge's theorem for convex functions]
\label{cor:berge}
Let $g:\mathbb R^n\to\mathbb R$ be a strictly convex function and let $\varphi:\mathbb R\rightrightarrows\mathbb R^n$ be a continuous correspondence that map into compact sets. Consider the function $h:\mathbb R\to \mathbb R^n$ given by
\[
    h(\lambda) = \arg\min_{\bfu'\in\varphi(\lambda)} g(\bfu')
\]
Then, $h$ is continuous. 
\end{Corollary}
\begin{proof}
Because $g$ is strictly convex, there is a unique minimizer $\bfu^\lambda$ of $g$ for each $\lambda\in\mathbb R$, so $h$ is well-defined. Furthermore, $h$ is upper hemicontinuous as a correspondence that maps real numbers $\lambda$ to singleton sets $\{h(\lambda)\}$ by Theorem \ref{thm:berge}, and any function $h$ that is upper hemicontinuous as a correspondence is continuous as a function. 
\end{proof}

\section{Equivalence of Group Sequential LASSO and Group Orthogonal Matching Pursuit}
\label{sec:sequential-lasso-vs-omp}

We will give our proof of Theorem \ref{thm:sequential-lasso-vs-omp} in this section.

\subsection{The dual problem}

Consider the Group Sequential LASSO objective:
\begin{equation}\label{eq:sequential-group-lasso}
\begin{aligned}
    \min_{\bfbeta\in\mathbb R^n} l(\bfbeta) + \lambda \sum_{i\in \overline S}\norm*{\bfbeta\vert_{T_i}}_2
\end{aligned}
\end{equation}
We will show that the dual of this problem is
\begin{equation}\label{eq:dual-sequential-group-lasso}
\begin{aligned}
    \max_{\bfu\in\mathbb R^n} -l^*(-\bfu) = -\min_{\bfu\in\mathbb R^n} l^*(-\bfu) \\
    \st \qquad \norm{\bfu\vert_{T_i}}_2 \leq \lambda ~\mbox{for each $i\in\overline S$} \\ \norm{\bfu\vert_{T_i}}_2 = 0 ~\mbox{for each $i\in S$}
\end{aligned}
\end{equation}

We write the objective of \eqref{eq:sequential-group-lasso} as a constrained optimization problem in the form of
\begin{align*}
    \min_{\bfz\in\mathbb R^n, \bfbeta\in\mathbb R^d} l(\bfz) + \lambda \sum_{i\in \overline S}\norm*{\bfbeta\vert_{T_i}}_2 \\
    \st \qquad \bfz = \bfbeta
\end{align*}
Then, the Lagrangian dual of this problem is
\begin{align*}
    \min_{\bfz\in\mathbb R^n, \bfbeta\in\mathbb R^n} \max_{\bfu\in\mathbb R^n}l(\bfz) + \lambda \sum_{i\in \overline S}\norm*{\bfbeta\vert_{T_i}}_2 + \bfu^\top(\bfz-\bfbeta)
\end{align*}
Furthermore, the objective of \eqref{eq:sequential-group-lasso} is convex and strictly feasible, so strong duality holds (see, e.g., Section 5.2.3 of \cite{BV2004}) and thus we may interchange the min and the max to obtain
\begin{align*}
    &\max_{\bfu\in\mathbb R^n}\min_{\bfz\in\mathbb R^n, \bfbeta\in\mathbb R^n} l(\bfz) + \lambda \sum_{i\in \overline S}\norm*{\bfbeta\vert_{T_i}}_2 + \bfu^\top(\bfz-\bfbeta) \\
    = ~&\max_{\bfu\in\mathbb R^n}\min_{\bfz\in\mathbb R^n} l(\bfz) + \bfu^\top\bfz + \min_{\bfbeta\in\mathbb R^n} \lambda \sum_{i\in \overline S}\norm*{\bfbeta\vert_{T_i}}_2 - \bfu^\top\bfbeta
\end{align*}
Now note that the first minimization over $\bfz\in\mathbb R^n$ gives exactly the Fenchel dual objective
\[
    \min_{\bfz\in\mathbb R^n} l(\bfz) + \bfu^\top\bfz = -\max_{\bfz\in\mathbb R^n} (-\bfu)^\top\bfz - l(\bfz) = -l^*(-\bfu).
\]
On the other hand, we show in the next lemma that the second minimization over $\bfbeta\in\mathbb R^n$ gives the constraints on the variables $\bfu$ given in \eqref{eq:dual-sequential-group-lasso}.

\begin{Lemma}
We have that
\[
    \inf_{\bfbeta\in\mathbb R^d} \lambda \sum_{i\in \overline S}\norm*{\bfbeta\vert_{T_i}}_2 - \bfu^\top\bfbeta = \begin{cases}
        0 & \text{if $\norm{\bfu\vert_{T_i}}_2 \leq \lambda$ for $i\in\overline S$ and $\norm{\bfu\vert_{T_i}}_2 = 0$ for $i\in S$} \\
        -\infty & \text{otherwise}
    \end{cases}
\]
\end{Lemma}
\begin{proof}
If $\norm{\bfu\vert_{T_i}}_2 > \lambda$ for some coordinate $i\in\overline S$, then we may choose $\bfbeta = \bfu\vert_{T_i}$ so that
\[
    \lambda \norm{\bfbeta\vert_{T_i}}_2 - \norm{\bfbeta\vert_{T_i}}_2^2 = \norm{\bfbeta\vert_{T_i}}_2(\lambda - \norm{\bfbeta\vert_{T_i}}_2) < 0
\]
so the objective can be made arbitrarily small by scaling. If $\norm{\bfu\vert_{T_i}}_2 > 0$ for some $i\in S$, then we may choose $\bfbeta = \bfu\vert_{T_i}$ so that
\[
    \lambda \sum_{i\in \overline S}\norm*{\bfbeta\vert_{T_i}}_2 - \norm{\bfu\vert_{T_i}}_2^2 = 0 - \norm{\bfu\vert_{T_i}}_2^2 < 0
\]
so the objective can be made arbitrarily small by scaling. Otherwise, we have that
\begin{align*}
    \bfu^\top\bfbeta &= \sum_{i\in \overline S}\bfu\vert_{T_i}^\top\bfbeta\vert_{T_i} && \text{since $\bfu\vert_{T_i} = 0$ for every $i\in S$} \\
    &\leq \sum_{i\in \overline S}\norm{\bfu\vert_{T_i}}_2 \norm{\bfbeta\vert_{T_i}}_2 && \text{Cauchy--Schwarz} \\
    &\leq \lambda \sum_{i\in \overline S} \norm{\bfbeta\vert_{T_i}}_2 && \text{since $\norm{\bfu\vert_{T_i}}_2 \leq \lambda$ for every $i\in\overline S$}.
\end{align*}
Thus,
\[
    \lambda \sum_{i\in \overline S}\norm*{\bfbeta\vert_{T_i}}_2 - \bfu^\top\bfbeta \geq 0
\]
and furthermore, this value can be achieved by $\bfbeta = 0$.
\end{proof}

\subsection{Selection of features}

We will use Berge's theorem (Theorem \ref{thm:berge}) to prove the following lemma, which characterizes the gradient of the optimal solution to the dual optimization problem given by \eqref{eq:dual-sequential-group-lasso}.

\begin{Lemma}
\label{lem:dual-solution-gradient}
Let $\lambda > 0$ and let $\bfu^\lambda$ be the minimizer of \eqref{eq:dual-sequential-group-lasso}. Let $\bfu^\infty$ be the minimizer of \eqref{eq:dual-sequential-group-lasso} without the constraint that $\norm{\bfu\vert_{T_i}}_2 \leq \lambda$ for every $i\in\overline S$. Define the threshold $\tau\coloneqq\max_{i\in\overline S}\norm*{\bfu^\infty\vert_{T_i}}_2$ and let $M^\tau\subseteq\overline S$ denote the corresponding set of indices $i\in\overline S$ that witnesses the max, that is,
\[
    M^\tau \coloneqq \braces*{i\in \overline S : \norm{\bfu^\infty\vert_{T_i}}_2 = \tau}.
\]
The following hold:
\begin{itemize}
    \item If $\lambda \geq \tau$, then $\nabla l^*(-\bfu^\lambda)\vert_{T_i}= 0$ for all $i\in\overline S$.
    \item If $\lambda = \tau - \eps$ for sufficiently small $\eps>0$, then $\nabla l^*(-\bfu^\lambda)\vert_{T_i} = 0$ for all $i\in\overline S \setminus M^\tau$ and $\nabla l^*(-\bfu^\lambda)\vert_{T_i} \neq 0$ for some $i\in M^\tau$.
\end{itemize}
\end{Lemma}
\begin{proof}
If $\lambda \geq \tau$, then the constraint $\max_{i\in\overline S}\norm*{\bfu^\lambda\vert_{T_i}}_2\leq\lambda$ can be removed without affecting the optimal solution, so $\bfu^\lambda = \bfu^\infty$. Then for the coordinates in $T_i$ for $i\in\overline S$, $\bfu^\infty$ is a minimizer for an unconstrained optimization problem, so the gradient is $0$ on these coordinates. This shows the first bullet point.

On the other hand, suppose that $\lambda = \tau - \eps$ for some small $\eps>0$. Then, $\bfu^\infty$ is outside the set $\braces{\bfu\in\mathbb R^n: \max_{i\in\overline S}\norm{\bfu\vert_{T_i}}_2 \leq \lambda}$. Now consider the function 
\[
    h(\lambda) = \max_{i\in\overline S\setminus M^\tau}\norm{\bfu^\lambda\vert_{T_i}}_2,
\]
i.e., the second largest value of $\norm{\bfu^\lambda\vert_{T_i}}_2$ after excluding the maximizers $i\in M^\tau$. Note that this function is continuous since $\lambda\mapsto \bfu^\lambda$ is continuous by Corollary \ref{cor:berge}. Furthermore, we have that $h(\tau) < \tau$, since the maximum in the definition of $h$ excludes the indices $M^\tau$. Let $\tau'$ satisfy $h(\tau) < \tau' < \tau$. Then, for all sufficiently small $\eps$, we have that $h(\tau-\eps) < \tau'$ by the continuity of $h$. For these $\eps$, we can remove the constraints of $\norm{\bfu\vert_{T_i}}_2 \leq \lambda = \tau-\eps$ for $i\in\overline S\setminus M^\tau$ without affecting the optimal solution $\bfu^\lambda$ in the optimization problem of \eqref{eq:dual-sequential-group-lasso}, so on the coordinates $T_i$ for $i\in\overline S\setminus M^\tau$, $\bfu^\lambda$ is an unconstrained minimizer and thus has zero gradient. On the other hand, for the coordinates $T_i$ for $i\in M^\tau$, $\bfu^\lambda$ cannot be the unconstrained minimizer and thus there must be some nonzero coordinate in the gradient due to the convexity of $l^*$.
\end{proof}

We can then show that Lemma \ref{lem:dual-solution-gradient} in fact characterizes the support of the optimal solution $\bfbeta^*$ by relating the primal and dual variables via the Fenchel--Young inequality (Theorem \ref{thm:fenchel-young}).

\begin{Lemma}[Primal vs dual variables]
We have that $-\bfu = \nabla l(\bfbeta)$ and $\bfbeta = \nabla l^*(-\bfu)$. 
\end{Lemma}
\begin{proof}
The primal variable $\bfz$ is related to the dual variable $\bfu$ via Fenchel dual, that is,
\[
    l^*(-\bfu) = (-\bfu)^\top\bfz - l(\bfz)
\]
Then by the tightness of the Fenchel--Young inequality (Theorem \ref{thm:fenchel-young}) for $l$, we have that $-\bfu = \nabla l(\bfz)$. Furthemore, by the conjugacy theorem (Theorem \ref{thm:fenchel-conjugacy}), we have that $(l^*)^* = l$, so $l^*(-\bfu) + (l^*)^*(\bfz) = (-\bfu)^\top\bfz$. Then by tightness of the Fenchel--Young inequality (Theorem \ref{thm:fenchel-young}) for $l^*$, we have that $\bfbeta = \bfz = \nabla l^*(-\bfu)$.
\end{proof}

Thus, by Lemma \ref{lem:dual-solution-gradient}, $\bfbeta^\lambda$ has a nonzero support on some group $T_i$ if and only if the group $T_i$ maximizes $\norm{\bfu^\infty\vert_{T_i}}_2 = \norm{\nabla l(\bfbeta^\infty)\vert_{T_i}}_2$. This is precisely the Group Orthogonal Matching Pursuit selection rule (see Line \ref{line:select-gradient} of Algorithm \ref{alg:group-omp}).

\bibliographystyle{alpha}
\bibliography{references}

\appendix

\section{Guarantees for Group Orthogonal Matching Pursuit}
\label{sec:group-omp}

In this section, we give guarantees for the Group OMP algorithm (Algorithm \ref{alg:group-omp}). Our analysis is similar to \cite{SSZ2010, LS2017, EKDN2018}. We first introduce the notion of restricted strong convexity and smoothness, generalized to the group setting.

\begin{Definition}[Restricted strong convexity and smoothness]
\label{def:rsc}
Let $l:\mathbb R^n\to\mathbb R$. Let $T_i\subseteq[n]$ for $i\in[t]$ form a partition of $[n]$. Then, $l$ is \emph{$\mu_s$-restricted strongly convex at group sparsity $s$} if for any $\bfbeta\in\mathbb R^n$ and $\bfDelta\in\mathbb R^n$ with $\norm{\bfDelta}_\group \leq s$,
\[
    l(\bfbeta + \bfDelta) - l(\bfbeta) - \angle*{\nabla l(\bfbeta), \bfDelta} \geq \frac{\mu_s}{2} \norm{\bfDelta}_2^2
\]
and \emph{$L_s$-restricted smooth at group sparsity $s$} if for any $\bfbeta\in\mathbb R^n$ and $\bfDelta\in\mathbb R^n$ with $\norm{\bfDelta}_\group \leq s$,
\[
    l(\bfbeta + \bfDelta) - l(\bfbeta) - \angle*{\nabla l(\bfbeta), \bfDelta} \leq \frac{L_s}{2}\norm{\bfDelta}_2^2.
\]
\end{Definition}

\begin{Lemma}[Smoothness]
\label{lem:omp-smoothness}
Let $l$ be $L_1$-restricted smooth at group sparsity $1$. Let $r\in[k']$ and let $\bfbeta^\infty$ and $i^*$ be defined as in Lines \ref{line:optimize-beta} and \ref{line:select-gradient} of Algorithm \ref{alg:group-omp} on the $r$-th iteration. Let $\bfbeta' \coloneqq \bfbeta^\infty + \bfDelta$ for $\bfDelta = - L_1^{-1}\nabla l(\bfbeta^\infty)\vert_{T_{i^*}}$. Then,
\[
    (2L_1)^{-1}\norm{\nabla l(\bfbeta^\infty)\vert_{T_{i^*}}}_2^2 \leq l(\bfbeta^\infty) - l(\bfbeta')
\]
\end{Lemma}
\begin{proof}
Note that $\bfDelta$ has group sparsity $1$. We then have that
\begin{align*}
    l(\bfbeta') - l(\bfbeta^\infty) &\leq \angle*{\nabla l(\bfbeta^\infty), \bfDelta} + \frac{L_1}{2}\norm{\bfDelta}_2^2 && \text{$L_1$-restricted smoothness} \\
    &= -L_1^{-1}\norm{\nabla l(\bfbeta^\infty)\vert_{T_{i^*}}}_2^2 + \frac12 L_1^{-1}\norm{\nabla l(\bfbeta^\infty)\vert_{T_{i^*}}}_2^2 \\
    &= -\frac12 L_1^{-1}\norm{\nabla l(\bfbeta^\infty)\vert_{T_{i^*}}}_2^2.
\end{align*}
Rearranging gives the desired result.
\end{proof}

\begin{Lemma}[Convexity]
\label{lem:omp-convexity}
Let $l$ be $\mu_{k+k'}$-restricted strongly convex at group sparsity $k+k'$. Let $r\in[k']$ and let $\bfbeta^\infty$ and $i^*$ be defined as in Lines \ref{line:optimize-beta} and \ref{line:select-gradient} of Algorithm \ref{alg:group-omp} on the $r$-th iteration. Let
\[
    \bfbeta^* \coloneqq \arg\min_{\bfbeta\in\mathbb R^n : \norm{\bfbeta}_\group \leq k} l(\bfbeta)
\]
Then,
\[
    \norm{\nabla l(\bfbeta^\infty)\vert_{T_{i^*}}}_2^2 \geq \frac{2\mu_{k+k'}}{k}\parens*{l(\bfbeta^\infty) - l(\bfbeta^*)}.
\]
\end{Lemma}
\begin{proof}
Let $U^*\subseteq[n]$ be the support of $\bfbeta^*$ and let $U\subseteq[n]$ be the support of $\bfbeta^\infty$. Note that $\norm*{\bfbeta^* - \bfbeta^\infty}_\group \leq k + k'$. Then,
\begin{align*}
    l(\bfbeta^*) - l(\bfbeta^\infty) &\geq \angle*{\nabla l(\bfbeta^\infty), \bfbeta^* - \bfbeta^\infty} + \frac{\mu_{k+k'}}2 \norm*{\bfbeta^* - \bfbeta^\infty}_2^2 \\
    &= \angle*{\nabla l(\bfbeta^\infty), (\bfbeta^* - \bfbeta^\infty)\vert_{U^*\setminus U}} + \frac{\mu_{k+k'}}2 \norm*{\bfbeta^* - \bfbeta^\infty}_2^2 && \nabla l(\bfbeta^\infty)\vert_U = \bfzero \\
    &\geq -\norm{\nabla l(\bfbeta^\infty)\vert_{U^*\setminus U}}_2\norm{(\bfbeta^* - \bfbeta^\infty)\vert_{U^*\setminus U}}_2 + \frac{\mu_{k+k'}}2 \norm*{(\bfbeta^* - \bfbeta^\infty)\vert_{U^*\setminus U}}_2^2 \\
    &\geq \min_x -\norm{\nabla l(\bfbeta^\infty)\vert_{U^*\setminus U}}_2 x + \frac{\mu_{k+k'}}2 x^2 \\
    & = -\frac{\norm{\nabla l(\bfbeta^\infty)\vert_{U^*\setminus U}}_2^2}{2\mu_{k+k'}}
\end{align*}
so
\[
    \norm{\nabla l(\bfbeta^\infty)\vert_{U^*\setminus U}}_2^2 \geq 2\mu_{k+k'}\parens*{l(\bfbeta^\infty) - l(\bfbeta^*)}.
\]
Now note that $U^*\setminus U$ is supported on at most $k$ groups, so by averaging, there exists some group $T_i$ outside of $U$ such that
\[
    \norm{\nabla l(\bfbeta^\infty)\vert_{T_i}}_2^2 \geq \frac{2\mu_{k+k'}}{k}\parens*{l(\bfbeta^\infty) - l(\bfbeta^*)}.\qedhere
\]
\end{proof}

Combining Lemmas \ref{lem:omp-smoothness} and \ref{lem:omp-convexity} leads to the following stepwise guarantee for Algorithm \ref{alg:group-omp}.

\begin{Lemma}
\label{lem:omp-stepwise}
Let $\bfbeta^{(r)}$ denote the value of $\bfbeta^\infty$ (Line \ref{line:optimize-beta}) after $r$ iterations of Algorithm \ref{alg:group-omp} with $\bfbeta^{(0)} = \bfzero$. Let 
\[
    \bfbeta^* \coloneqq \arg\min_{\bfbeta\in\mathbb R^n : \norm{\bfbeta}_\group \leq k} l(\bfbeta)
\]
Then,
\[
    l(\bfbeta^{(r)}) - l(\bfbeta^*) \leq \exp\parens*{-\frac{r}{k}\frac{\mu_{k+k'}}{L_1}}\parens*{l(\bfbeta^{(0)}) - l(\bfbeta^*)}
\]
\end{Lemma}
\begin{proof}
By Lemmas \ref{lem:omp-smoothness} and \ref{lem:omp-convexity}, we have that
\[
    l(\bfbeta^{(r)}) - l(\bfbeta^{(r+1)}) \geq (2L_1)^{-1}\norm{\nabla l(\bfbeta^{(r)})\vert_{T_{i^*}}}_2^2 \geq \frac1k\frac{\mu_{k+k'}}{L_1} \parens*{l(\bfbeta^{(r)}) - l(\bfbeta^*)}
\]
so
\begin{align*}
    l(\bfbeta^{(r+1)}) - l(\bfbeta^*) &= l(\bfbeta^{(r)}) - l(\bfbeta^*) - \parens*{l(\bfbeta^{(r)}) - l(\bfbeta^{(r+1)})} \\
    &\leq l(\bfbeta^{(r)}) - l(\bfbeta^*) - \frac1k \frac{\mu_{k+k'}}{L_1} \parens*{l(\bfbeta^{(r)}) - l(\bfbeta^*)} \\
    &= \parens*{1 - \frac1k \frac{\mu_{k+k'}}{L_1}} \parens*{l(\bfbeta^{(r)}) - l(\bfbeta^*)} \\
    &\leq \exp\parens*{- \frac1k \frac{\mu_{k+k'}}{L_1}}\parens*{l(\bfbeta^{(r)}) - l(\bfbeta^*)}
\end{align*}
Applying the above inductively proves the claim.
\end{proof}

As a result of Lemma \ref{lem:omp-stepwise}, we obtain two guarantees for Algorithm \ref{alg:group-omp}, one for exact $k$-group-sparse solutions with large approximation and one for bicriteria sparsity with $\eps$ additive error.

\CorSparseOMP
\begin{proof}
After $k$ iterations, we have by Lemma \ref{lem:omp-stepwise}
applied for $k'=k$ that
\[
    l(\bfbeta^{(k)}) - l(\bfbeta^*) = l(\bfbeta^{(k)}) - l(\bfbeta^{(0)}) + l(\bfbeta^{(0)})- l(\bfbeta^*) \leq \exp\parens*{-\frac{\mu_{2k}}{L_1}}\parens*{l(\bfbeta^{(0)}) - l(\bfbeta^*)}
\]
which rearranges to
\[
    l(\bfbeta^{(0)}) - l(\bfbeta^{(k)}) \geq \parens*{1-\exp\parens*{-\frac{\mu_{2k}}{L_1}}}\parens*{l(\bfbeta^{(0)}) - l(\bfbeta^*)}
\]
\end{proof}

\CorBicriteriaOMP
\begin{proof}
This follows immediately from the bound of Lemma \ref{lem:omp-stepwise} and rearranging.
\end{proof}

\subsection{Group OMP with Replacement}

In this section, we give guarantees for the Group OMP with Replacement algorithm (Algorithm \ref{alg:group-ompr}), which is an improvement to Group OMP that can achieve a sparsity bound that is independent of the accuracy parameter $\eps$ \cite{AS2020}.

\begin{algorithm}
\caption{Group Orthogonal Matching Pursuit with Replacement.}
\label{alg:group-ompr}
\begin{algorithmic}[1]
  \Function{GroupOMPR}{objective $l$, sparsity $k$, initial sparsity $k'$, iterations $R$}
  \State Initialize $S^0\subseteq[n]$ with $|S^0|=k'$, e.g. using Algorithm~\ref{alg:group-omp}.
  \For{$r=0$ to $R-1$}
    \State Let
    \[
        \bfbeta^\infty \coloneqq \arg\min_{\substack{\bfbeta\in\mathbb R^n \\ \forall i\in\overline S^r, \bfbeta\vert_{T_i} = 0}} l(\bfbeta)
    \] \label{line:optimize-beta-ompr}
    \State Let $i^*\in\overline S^r$ be such that $\norm{\nabla l(\bfbeta^\infty)\vert_{T_{i^*}}}_2^2 = \max_{i\in\overline S^r}\norm{\nabla l(\bfbeta^\infty)\vert_{T_i}}_2^2$ \label{line:select-gradient-ompr}
    \State Let $j^*\in S^r$ be such that $\norm{\bfbeta^\infty\vert_{T_{j^*}}}_2^2 = \min_{j\in S}\norm{\bfbeta^\infty\vert_{T_j}}_2^2$ \label{line:remove-group-ompr}
    \State Update $S^{r+1} \gets S^r \cup \{i^*\} \setminus \{j^*\}$
  \EndFor
  \State \Return $S^r$, $r\in[R]$, that minimizes \[\min_{\substack{\bfbeta\in\mathbb R^n \\ \forall i\in\overline S^r, \bfbeta\vert_{T_i} = 0}} l(\bfbeta)\]
  \EndFunction
\end{algorithmic}
\end{algorithm}

\begin{Lemma}[Smoothness]
\label{lem:ompr-smoothness}
Let $l$ be $L_2$-restricted smooth at group sparsity $2$. Let $r\in[k']$ and let $\bfbeta^\infty$, $i^*$, $j^*$ be defined as in Lines \ref{line:optimize-beta-ompr}, \ref{line:select-gradient-ompr} and \ref{line:remove-group-ompr} of Algorithm \ref{alg:group-ompr} on the $r$-th iteration. Let $\bfbeta' \coloneqq \bfbeta^\infty + \bfDelta$ for $\bfDelta = - L_2^{-1}\nabla l(\bfbeta^\infty)\vert_{T_{i^*}} - \bfbeta^\infty\vert_{T_{j^*}}$. Then,
\[
    (2L_2)^{-1}\norm{\nabla l(\bfbeta^\infty)\vert_{T_{i^*}}}_2^2 
    - (1/2) L_2 \norm{\bfbeta^\infty\vert_{T_{j^*}}}_2^2 \leq l(\bfbeta^\infty) - l(\bfbeta')
\]
\end{Lemma}
\begin{proof}
Note that $\bfDelta$ has group sparsity $2$. We then have that
\begin{align*}
    l(\bfbeta') - l(\bfbeta^\infty) &\leq \angle*{\nabla l(\bfbeta^\infty), \bfDelta} + \frac{L_2}{2}\norm{\bfDelta}_2^2 && \text{$L_2$-restricted smoothness} \\
    & =
    -L_2^{-1}\norm{\nabla l(\bfbeta^\infty)\vert_{T_{i^*}}}_2^2
    + \frac12 L_2^{-1}\norm{\nabla l(\bfbeta^\infty)\vert_{T_{i^*}}}_2^2 
    + \frac12 L_2 \norm{\bfbeta^\infty\vert_{T_{j^*}}}_2^2
    && (\norm{\nabla l(\bfbeta^\infty)\vert_{T_{j^*}}}_2^2=0) \\
    &= -\frac12 L_2^{-1}\norm{\nabla l(\bfbeta^\infty)\vert_{T_{i^*}}}_2^2 + \frac12 L_2\norm{\bfbeta^\infty\vert_{T_{j^*}}}_2^2.
\end{align*}
Rearranging gives the desired result.
\end{proof}
\begin{Lemma}[Convexity]
\label{lem:ompr-convexity}
Let $l$ be $\mu_{k+k'}$-restricted strongly convex at group sparsity $k+k'$. Let $r\in[k']$ and let $\bfbeta^\infty$, $i^*$, $j^*$ be defined as in Lines \ref{line:optimize-beta-ompr}, \ref{line:select-gradient-ompr} and \ref{line:remove-group-ompr} of Algorithm \ref{alg:group-ompr} on the $r$-th iteration. Let
\[
    \bfbeta^* \coloneqq \arg\min_{\bfbeta\in\mathbb R^n : \norm{\bfbeta}_\group \leq k} l(\bfbeta)
\]
Then,
\[
    \norm{\nabla l(\bfbeta^\infty)\vert_{T_{i^*}}}_2^2 \geq \frac{2\mu_{k+k'}}{k}\parens*{l(\bfbeta^\infty) - l(\bfbeta^*)}
    + \frac{(k'-k)\mu_{k+k'}^2}{k} \norm{\bfbeta^\infty\vert_{T_{j^*}}}_2^2.
\]
\end{Lemma}
\begin{proof}
Let $U^*\subseteq[n]$ be the support of $\bfbeta^*$ and let $U\subseteq[n]$ be the support of $\bfbeta^\infty$. Note that $\norm*{\bfbeta^* - \bfbeta^\infty}_\group \leq k + k'$. Then,
\begin{align*}
    & l(\bfbeta^*) - l(\bfbeta^\infty) \\
    &\geq \angle*{\nabla l(\bfbeta^\infty), \bfbeta^* - \bfbeta^\infty} + \frac{\mu_{k+k'}}2 \norm*{\bfbeta^* - \bfbeta^\infty}_2^2 \\
    &= \angle*{\nabla l(\bfbeta^\infty), (\bfbeta^* - \bfbeta^\infty)\vert_{U^*\setminus U}} + \frac{\mu_{k+k'}}2 \norm*{\bfbeta^* - \bfbeta^\infty}_2^2 \\
    &\geq -\norm{\nabla l(\bfbeta^\infty)\vert_{U^*\setminus U}}_2\norm{(\bfbeta^* - \bfbeta^\infty)\vert_{U^*\setminus U}}_2 + \frac{\mu_{k+k'}}2 \norm*{(\bfbeta^* - \bfbeta^\infty)\vert_{U^*\setminus U}}_2^2 
    + \frac{\mu_{k+k'}}2 \norm*{(\bfbeta^* - \bfbeta^\infty)\vert_{U\setminus U^*}}_2^2 \\
    &\geq \min_x -\norm{\nabla l(\bfbeta^\infty)\vert_{U^*\setminus U}}_2 x + \frac{\mu_{k+k'}}2 x^2 
    + \frac{\mu_{k+k'}}2 \norm*{\bfbeta^\infty\vert_{U\setminus U^*}}_2^2\\
    & = -\frac{\norm{\nabla l(\bfbeta^\infty)\vert_{U^*\setminus U}}_2^2}{2\mu_{k+k'}}
    + \frac{\mu_{k+k'}}2 \norm*{\bfbeta^\infty\vert_{U\setminus U^*}}_2^2
\end{align*}
so
\[
    \norm{\nabla l(\bfbeta^\infty)\vert_{U^*\setminus U}}_2^2 \geq 2\mu_{k+k'}\parens*{l(\bfbeta^\infty) - l(\bfbeta^*)}
    + \mu_{k+k'}^2 \norm*{\bfbeta^\infty\vert_{U\setminus U^*}}_2^2.
\]
Now note that $U^*\setminus U$ is supported on at most $k$ groups, so by averaging, there exists some group $T_i$ outside of $U$ such that
\begin{align*}
    \norm{\nabla l(\bfbeta^\infty)\vert_{T_i}}_2^2 
    & \geq \frac{2\mu_{k+k'}}{k}\parens*{l(\bfbeta^\infty) - l(\bfbeta^*)}
    + \frac{\mu_{k+k'}^2}{k} \norm*{\bfbeta^\infty\vert_{U\setminus U^*}}_2^2\\
    & \geq \frac{2\mu_{k+k'}}{k}\parens*{l(\bfbeta^\infty) - l(\bfbeta^*)}
    + \frac{(k'-k)\mu_{k+k'}^2}{k} \norm{\bfbeta^\infty\vert_{T_{j^*}}}_2^2.\qedhere
\end{align*}
\end{proof}

\begin{Lemma}
\label{lem:ompr-stepwise}
Let $\bfbeta^{(r)}$ denote the value of $\bfbeta^\infty$ (Line \ref{line:optimize-beta-ompr}) after $r$ iterations of Algorithm \ref{alg:group-ompr} with $\bfbeta^{(0)} = \bfzero$ and $|S^0| = k' \geq k\parens*{\frac{L_2^2}{\mu_{k+k'}^2} + 1}$. Let 
\[
    \bfbeta^* \coloneqq \arg\min_{\bfbeta\in\mathbb R^n : \norm{\bfbeta}_\group \leq k} l(\bfbeta)
\]
Then,
\[
    l(\bfbeta^{(r)}) - l(\bfbeta^*) \leq \exp\parens*{-\frac{r}{k}\frac{\mu_{k+k'}}{L_2}}\parens*{l(\bfbeta^{(0)}) - l(\bfbeta^*)}
\]
\end{Lemma}
\begin{proof}
By Lemmas \ref{lem:ompr-smoothness} and \ref{lem:ompr-convexity}, we have that
\begin{align*}
    l(\bfbeta^{(r)}) - l(\bfbeta^{(r+1)}) 
    & \geq (2L_2)^{-1}\norm{\nabla l(\bfbeta^{(r)})\vert_{T_{i^*}}}_2^2 - (1/2)L_2\left\|\bfbeta^\infty\vert_{T_{j^*}}\right\|_2^2\\
    & \geq \frac1k\frac{\mu_{k+k'}}{L_2} \parens*{l(\bfbeta^{(r)}) - l(\bfbeta^*)}
    + \frac12 \parens*{\frac{(k'-k)\mu_{k+k'}^2}{kL_2}- L_2}\left\|\bfbeta^\infty\vert_{T_{j^*}}\right\|_2^2\\
    & \geq \frac1k\frac{\mu_{k+k'}}{L_2} \parens*{l(\bfbeta^{(r)}) - l(\bfbeta^*)},
\end{align*}
as long as $k' \geq k\parens*{L_2^2 / \mu_{k+k'}^2 + 1}$.
So,
\begin{align*}
    l(\bfbeta^{(r+1)}) - l(\bfbeta^*) &= l(\bfbeta^{(r)}) - l(\bfbeta^*) - \parens*{l(\bfbeta^{(r)}) - l(\bfbeta^{(r+1)})} \\
    &\leq l(\bfbeta^{(r)}) - l(\bfbeta^*) - \frac1k \frac{\mu_{k+k'}}{L_2} \parens*{l(\bfbeta^{(r)}) - l(\bfbeta^*)} \\
    &= \parens*{1 - \frac1k \frac{\mu_{k+k'}}{L_2}} \parens*{l(\bfbeta^{(r)}) - l(\bfbeta^*)} \\
    &\leq \exp\parens*{- \frac1k \frac{\mu_{k+k'}}{L_2}}\parens*{l(\bfbeta^{(r)}) - l(\bfbeta^*)}
\end{align*}
Applying the above inductively proves the claim.
\end{proof}

\CorBicriteriaOMPR
\begin{proof}
This follows immediately from the bound of Lemma \ref{lem:ompr-stepwise} and rearranging.
\end{proof}

\section{Equivalence of Group Sequential Attention and Group Sequential LASSO}

We generalize a result of \cite{Hof2017} to the group setting, which allows us to translate guarantees for Group Sequential LASSO (Algorithm \ref{alg:group-sequential-lasso}) to Group Sequential Attention (Algorithm \ref{alg:group-sequential-attention}). 

\begin{Lemma}
\label{lem:group-hoff}
Let $l:\mathbb R^n\to\mathbb R$ and $\lambda>0$. Let $T_i\subseteq[n]$ for $i\in[t]$ form a partition of $[n]$. Let $S\subseteq[t]$. Then,
\[
    \inf_{\bfbeta\in\mathbb R^n} l(\bfbeta) + \lambda \sum_{i\in\overline S}\norm{\bfbeta\vert_{T_i}}_2 = \inf_{\bfw\in\mathbb R^t, \bfbeta\in\mathbb R^n} l(\bfbeta_{\bfw}) + \frac{\lambda}{2}\parens*{\norm{\bfw\vert_{\overline S}}_2^2 + \sum_{i\in\overline S}\norm{\bfbeta\vert_{T_i}}_2^2 }
\]
where $\bfbeta_{\bfw}\in\mathbb R^n$ is the vector such that $\bfbeta_{\bfw}\vert_{T_i} \coloneqq \bfw_i \cdot \bfbeta\vert_{T_i}$.
\end{Lemma}
\begin{proof}
We have that
\begin{align*}
    \inf_{\bfw\in\mathbb R^t, \bfbeta\in\mathbb R^n} l(\bfbeta_{\bfw}) + \frac{\lambda}{2}\parens*{\norm{\bfw\vert_{\overline S}}_2^2 + \sum_{i\in\overline S}\norm{\bfbeta\vert_{T_i}}_2^2 } &= \inf_{\bfw\in\mathbb R^t, \bfu\in\mathbb R^n} l(\bfu) + \frac{\lambda}{2}\sum_{i\in\overline S}\bfw_i^2 + \frac{\norm{\bfu\vert_{T_i}}_2^2}{\bfw_i^2} 
\end{align*}
Now note that for each $i\in \overline S$, we have that
\[
    \bfw_i^2 + \frac{\norm{\bfu\vert_{T_i}}_2^2}{\bfw_i^2} \geq 2\norm{\bfu\vert_{T_i}}_2
\]
with equality if and only if $\bfw_i^2 = \norm{\bfu\vert_{T_i}}_2$ by tightness of the AM-GM inequality.
\end{proof}

\end{document}